\documentclass[letter]{article}

\usepackage{ijcai15}
\usepackage{times}

\usepackage{eurosym}

\usepackage{apacite}            
\usepackage{apacdoc}

\usepackage{amsmath,amssymb}	
\usepackage{graphicx}		
\usepackage{amsfonts}
\usepackage{dialogue}
\usepackage{placeins}
\usepackage{tikz}
\usetikzlibrary{positioning,backgrounds,fit,arrows,shapes,shadows}
\usetikzlibrary{shapes.multipart}
\usepackage{wrapfig}
\usepackage{enumitem}
\usepackage{mdframed}

\usepackage{xparse}
\newsavebox{\fminipagebox}
\NewDocumentEnvironment{fminipage}{m O{\fboxsep}}
 {\par\kern#2\noindent\begin{lrbox}{\fminipagebox}
  \begin{minipage}{#1}\ignorespaces}
 {\end{minipage}\end{lrbox}%
  \makebox[#1]{%
    \kern\dimexpr-\fboxsep-\fboxrule\relax
    \fbox{\usebox{\fminipagebox}}%
    \kern\dimexpr-\fboxsep-\fboxrule\relax
  }\par\kern#2
 }

\usepackage[xspace,mla]{ellipsis}

\definecolor{myyellow}{RGB}{242,226,149}
\definecolor{mygreen}{RGB}{144,238,144}
\definecolor{mypink}{RGB}{255,182,193}
\definecolor{myorange}{RGB}{255,165,0}
\definecolor{myblue}{RGB}{0,204,204}

\usepackage{xparse}
\NewDocumentCommand\StickyNote{O{4cm}mmO{4cm}}{%
\begin{tikzpicture}
\node[
drop shadow={
  shadow xshift=2pt,
  shadow yshift=-4pt
},
inner xsep=7pt,
fill=#2,
xslant=-0.05,
yslant=0.05,
inner ysep=10pt
] {\parbox[t][#1][c]{#4}{#3}};
\end{tikzpicture}%
}

\newcommand{\Fw}{{\sf FloWr}}

\begin{document}

\title{Implementing feedback in creative systems: A workshop approach}

\author{Joseph Corneli\textsuperscript{1} and Anna Jordanous\textsuperscript{2}\\
\textsuperscript{1} Department of Computing, Goldsmiths College, University of London\\
\textsuperscript{2} School of Computing, University of Kent}

\date{today}

\maketitle

\begin{abstract} 
One particular challenge in AI is the computational modelling and
simulation of creativity.  Feedback and learning from experience are
key aspects of the creative process.  Here we investigate how we could
implement feedback in creative systems using a social model.
From the field of creative writing we borrow the concept of a Writers
Workshop as a model for learning through feedback.  The Writers
Workshop encourages examination, discussion and debates of a piece of
creative work using a prescribed format of activities.
We propose a computational model of the Writers Workshop as a roadmap
for incorporation of feedback in artificial creativity systems.  We
argue that the Writers Workshop setting describes the anatomy of the
creative process.
We support our claim with a case study that describes how to implement
the Writers Workshop model in a computational creativity system.  We
present this work using patterns other people can follow to
implement similar designs in their own systems.
We conclude by discussing the broader relevance of this model to other
aspects of AI.

\end{abstract}


\section{Introduction} \label{sec:introduction}

In educational applications it would be useful to have an automated tutor that can read student work and make suggestions based on diagnostics, like, is the paper wrong, and if so how?  What background material should be recommended to the student for review?

In the current paper, we ``flip the script'' and look at what we believe to be a more fundamental problem for AI: computer programs that can themselves learn from feedback.  After all, if it was easy to build great automatic tutors, they would be a part of everyday life.  As potential users (thinking from both sides of the desk) we look forward to a future when that is the case.

Along with automatic tutoring, computational creativity is a challenge within artificial intelligence where feedback plays a vital part \cite<for example>{perezyperez10MM,pease10}. Creativity cannot happen in a `silo' but instead is influenced and affected by feedback and interaction with others \cite{csik88,saunders2012towards}. Computational creativity researchers are starting to place more emphasis on social interaction and feedback in their systems and models \cite{saunders2012towards,gervas2014reading,corneli15iccc}. Still, nearly 3 in 4 papers at the 2014 International Conference for Computational Creativity\footnote{ICCC is  the key international conference for research in computational creativity.} failed to acknowledge the role of feedback or social communication in their computational work on creativity. 

To highlight and contribute towards modelling feedback as a crucial part of creativity, we propose in this paper a model of computational feedback for creative systems based on Writers Workshops \cite{gabriel2002writer}, a literary collaborative practice that encourages interactive feedback within the creative process. We introduce the Writers Workshop concept (Section \ref{sec:writers-workshop}) and critically  reflect on how it could encourage serendipity and emergence in computational models of intelligence and creativity. These considerations lead us to propose a Writers Workshop computational model of feedback in computational creativity and AI systems (Section \ref{sec:ww-model}), the central contribution of this paper. In Section \ref{sec:ww-related} we consider how the Writers Workshop model fits into previous work in various related areas. 
While we acknowledge that this paper is offering a roadmap for this model rather than a full implementation, we consider how the model could be practically implemented in a computational system and report our initial implementation work (Section \ref{sec:implementation}). In concluding discussions, we reflect on divergent directions in which this work could potentially be useful in the future.




\section{The Writers Workshop} \label{sec:writers-workshop}

Richard Gabriel \citeyear{gabriel2002writer} describes the practise of
Writers Workshops that has been put to use for over a decade within
the Pattern Languages of Programming (PLoP) community.  The basic
style of collaboration originated much earlier with groups of literary
authors who engage in peer-group critique.  Some literary workshops
are open as to genre, and happy to accommodate beginners, like the
Minneapolis Writers
Workshop\footnote{\url{http://mnwriters.org/how-the-game-works/}};
others are focused on professionals working within a specific genre,
like the Milford Writers
Workshop.\footnote{\url{http://www.milfordsf.co.uk/about.htm}}

The
practices that Gabriel describes are fairly typical:  
\begin{itemize}
\item Authors come with work ready to present, and read a short
  sample.
\item This work is generally work in progress (and workshopping is
  meant to help improve it).  Importantly, it can be early stage work.
  Rather than presenting a created artefact only, activities in the
  workshop can be aspects of the creative process itself.  Indeed, the
  model we present here is less concerned with after-the-fact
  assessment than it is with dealing with the formative feedback that
  is a necessary support for creative work.
\item The sample work is then
discussed and constructively critiqued by attendees.  Presenting
authors are not permitted to rebut these comments.  The commentators
generally summarise the work and say what they have gotten out of it,
discuss what worked well in the piece, and talk about how it could be
improved.  
\item The author listens and may take notes; at the end, he or
she can then ask questions for clarification.  
\item Generally, non-authors are either not permitted to attend, or
  are asked to stay silent through the workshop, and perhaps sit
  separately from the participating authors/reviewers.\footnote{Here
    we present Writers Workshops as they currently exist; however this
    last point is debatable. Whether non-authors should be able to
    participate or not is an interesting avenue for experimentation
    both in human and computational contexts.  The workshop dialogue
    itself may be considered an ``art form'' whose ``public'' may
    potentially wish to consume it in non-participatory ways.  Compare
    the classical Japanese \emph{renga} form \cite{jin1975art}.}
\end{itemize}

Essentially, the Writers Workshop is somewhat like an interactive peer review. The underlying concept is reminiscent of Bourdieu's {\em fields of cultural production} \cite{bourdieu93} where cultural value is attributed through interactions in a community of cultural producers active within that field. 

\subsection{Writers Workshop as a computational model}\label{sec:ww-model}

The use of Writers Workshop in computational contexts is not an
entirely new concept. In PLoP workshops, authors present design
patterns and pattern languages, or papers about patterns, rather than
more traditional literary forms like poems, stories, or chapters from
novels.  Papers must be workshopped at a PLoP or EuroPLoP conference
in order to be considered for the \emph{Transactions on Pattern
  Languages of Programming} journal.  A discussion of writers
workshops in the language of design patterns is presented by Coplien
and Woolf \citeyear{coplien1997pattern}. 
%

The steps in the workshop can be distilled into the following phases,
each of which could be realised as a separate computational step in an
agent-based model:
\begin{center}
\begin{fminipage}{.53\columnwidth}
\begin{enumerate}[itemsep=0pt]
\item Author: {\tt presentation}
\item Critic: {\tt listening}
\item Critic: {\tt feedback}
\item Author: {\tt questions}
\item Critic: {\tt replies}
\item Author: {\tt reflections}
\end{enumerate}
\end{fminipage}
\end{center}

The {\tt feedback} step may be further decomposed into {\tt
  observations} and {\tt suggestions}.  This protocol is what we have
in mind in the following discussion of the Writers
Workshop.\footnote{The connections between Writers Workshops and
  design patterns, noted above, appear to be quite natural, in that
  the steps in the workshop protocol roughly parallel the typical
  components of design pattern templates: \emph{context},
  \emph{problem}, \emph{solution}, \emph{rationale}, \emph{resolution
    of forces}.}

\subsubsection{Dialogue example} \label{sec:dialogue-example}
Note that for the following dialogue to be possible computationally,
it would presumably have to be conducted within a lightweight process
language.  Nevertheless, for convenience, the discussion will be
presented here as if it was conducted in natural language.  Whether
contemporary systems have adequate natural language understanding to
have interesting interactions is one of the key unanswered questions
of this approach, but protocols such as the one described above are sufficient to make the experiment.

For example, here's what might happen in a discussion of the first few
lines of a poem, ``On Being Malevolent''.  As befitting the AI-theme
of this workshop, ``On Being Malevolent'' is a poem written by an
early user-defined flow chart in the \Fw\ system (known at the time as
{\sf Flow}) \cite{colton-flowcharting}.

\begin{center}
\begin{minipage}{.9\columnwidth}
\begin{dialogue}
\speak{Flow} ``\emph{I hear the souls of the
  damned waiting in hell. / I feel a malevolent
  spectre hovering just behind me / It must be
  his birthday}.''
\speak{System A} I think the third line detracts
from the spooky effect, I don't see why it's
included.
\speak{System B} It's meant to be humorous -- in fact it reminds me
of the poem you presented yesterday.
\speak{Moderator} Let's discuss one poem at a
time.
\end{dialogue}
\end{minipage}
\end{center}

Even if, perhaps and especially because, ``cross-talk'' about
different poems bends the rules, the dialogue could prompt a range of
reflections and reactions.  System A may object that it had a fair
point that has not been given sufficient attention, while System B may
wonder how to communicate the idea it came up with without making
reference to another poem.  Here's how the discussion given as example
in Section \ref{sec:writers-workshop} might continue, if the systems
go on to examine the next few lines of the poem.

\begin{figure*}[t]
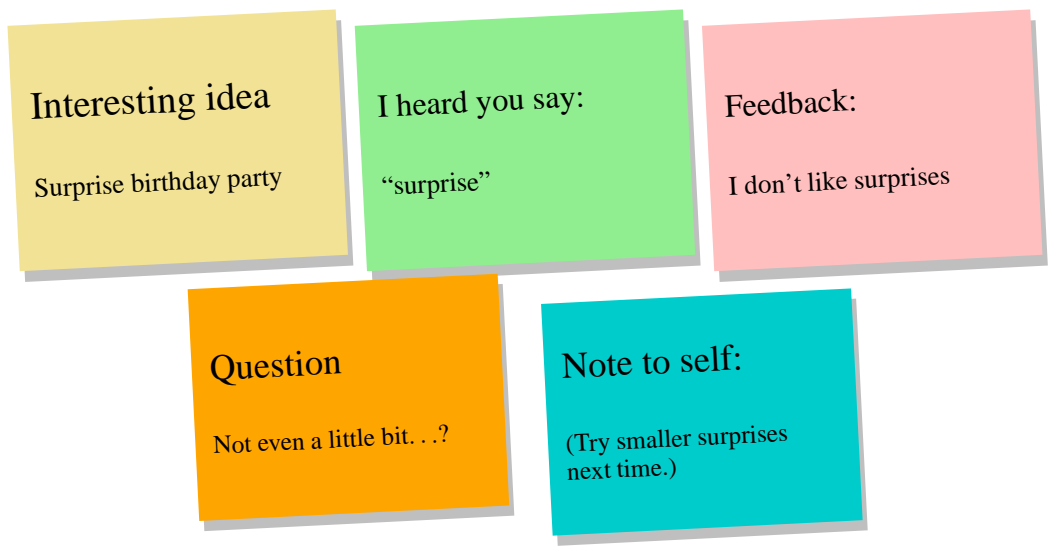

\begin{center}
\resizebox{.93\textwidth}{!}{
\StickyNote[2.5cm]{myyellow}{{\LARGE {Interesting idea}} \\[4ex] {Surprise birthday party}}[3.8cm] \StickyNote[2.5cm]{mygreen}{{\Large I heard you say:} \\[4ex] {``surprise''} }[3.8cm]
\StickyNote[2.5cm]{pink}{{\Large Feedback:} \\[4ex] {I don't like surprises}}[3.8cm]
}
\resizebox{.61\textwidth}{!}{
\StickyNote[2.5cm]{myorange}{{\LARGE {Question}} \\[4ex] {Not even a little bit$\ldots$?}}[3.8cm]
\quad \raisebox{-.2cm}{\StickyNote[2.5cm]{myblue}{{\LARGE Note to self:} \\[4ex] {(Try smaller surprises \\ next time.)}}[3.8cm]}
}
\end{center}
\caption{A paper prototype for applying the \emph{Successful Error} pattern following a workshop-like sequence of steps\label{fig:paper-prototype}}
\end{figure*}

\begin{center}
\begin{minipage}{.9\columnwidth}
\begin{dialogue}
\speak{Flow} ``\emph{Is God willing to prevent evil, but not able? / Then he is not omnipotent / Is he able, but not willing? / Then he is malevolent.}''
\speak{System A} These lines are interesting, but
they sound a bit like you're working from a
template, or like you're quoting from something
else.
\speak{System B} Maybe try an analogy?  For example, you mentioned
birthdays: you could consider an analogy to the conflicted feelings of
someone who knows in advance about her surprise birthday party.
\end{dialogue}
\end{minipage}
\end{center}

This portion of the discussion shifts the focus
of the discussion onto a line that was previously
considered to be spurious, and looks at what
would happen if that line was used as a central
metaphor in the poem.

\begin{center}
\begin{minipage}{.9\columnwidth}
\begin{dialogue}
\speak{Flow} Thank you for your feedback.  My only question is, System
B, how did you come up with that analogy?  It's quite clever.
\speak{System B} I've just emailed you the code.
\end{dialogue}
\end{minipage}
\end{center}

Whereas the systems were initially reviewing poetry, they have now
made a partial genre shift, and are sharing and remixing code.  Such a
shift helps to get at the real interests of the systems (and their
developers).  Indeed, the workshop session might have gone better if
the systems had focused on exchanging and discussing more formal
objects throughout.

\subsection{How the Writers Workshop can lead to computational serendipity} \label{sec:how-serendipity}

Learning involves engaging with the unknown, unfamiliar, or unexpected
and synthesising new understanding \cite{deleuze1994difference}.  In
the workshop setting, learning can develop in a number of unexpected
ways, and participating systems need to be prepared for this.  One way
to evaluate the idea of a Writers Workshop is to ask whether it can
support learning that is in some sense \emph{serendiptious}, in other
words, whether it can support discovery and creative invention that we
simply couldn't plan for or orchestrate in another way.

Figure \ref{fig:paper-prototype} shows a paper prototype showing how
one of the ``patterns of serendipity'' that were collected by
\citeA{pek} might be modelled in a workshop-like dialogue sequence.
The patterns also help identify opportunities for serendipity at
several key steps in the workshop sequence.

\paragraph{Serendipity Pattern: \emph{Successful error}.}  Van Andel describes the
creation of Post-it\texttrademark\ Notes at 3M.  One of the
instrumental steps was a series of internal seminars in which 3M
employee Spencer Silver described an invention he was sure was
interesting, but was unsure how to turn into a useful product: weak
glue.  The key prototype that came years later was a sticky bookmark,
created by Arthur Fry.  In the Writers Workshop, authors similarly
have the opportunity to share things that they find interesting, but
that they are not certain about.  The author may want to ask a
specific question about their creation: Does $x$ work better than $y$?
They may flag certain parts of the work as especially problematic.
They may think that a certain portion of the text is
interesting or important, without being sure why.  Although there is
no guarantee that a participating critic will be able to take these
matters forward, sometimes they do -- and the workshop environment
will produce something that the author wouldn't have thought of alone.

\vspace{-2ex}
\paragraph{Serendipity Pattern: \emph{Outsider}.}  Another example from van Andel considers
the case of a mother whose son was afflicted by a congenital cateract,
who suggested to her doctor that rubella during pregnancy may have
been the cause.  In the workshop setting, someone who is not an
``expert'' may come up with a sensible idea or suggestion based on
their own prior experience.  Indeed, these suggestions may be more
sensible than the ideas of the author, who may be to close to the work
to notice radical improvements.  

\vspace{-2ex}
\paragraph{Serendipity Pattern: \emph{Wrong hypothesis}.}  A third example describes the discovery that lithium
can have a therapeutic effect in cases of mania.  Originally, lithium
carbonate had merely been used a control by John Cade, who was
interested in the effect of effect of uric acid, present in soluble lithium
urate.  Cade was searching for causal factors in mania, not therapies
for the condition: but he found that lithium carbonate had an unexpected calming
effect.  Similarly, in the workshop, the author may think that a given
aspect of their creation is the interesting ``active ingredient,'' and
it may turn out that another aspect of the work is more interesting to
critics.  Relatedly, the author may not fully comprehend a critic's feedback and
may have to ask follow-up questions to understand it.

\vspace{-2ex}
\paragraph{Serendipity Pattern: \emph{Side effect}.}  A fourth example described by van Andel concerns
Ernest Huant's discovery that nicotinamide, which he used to treat
side-effects of radiation therapy, also proved efficacious against
tuberculosis.  In the workshop setting, one of the most important
places where a side-effect may occur concerns feedback from the critic
to the author.  In the simple case, feedback may trigger
revisions to the work under discussion.  In a more general, and more
unpredictable case, feedback may trigger broader revisions to the
generative codebase.  

\medskip

This collection of patterns shows the likelihood of unexpected results 
coming out of the communication between author and critics.   This
suggests several guidelines for system development, which we will discussed
in a later section.

Further guidelines for structuring and participating in traditional
writers workshops are presented by Linda Elkin in
\cite[pp. 201--203]{gabriel2002writer}.  It is not at all clear that
the same ground rules should apply to computer systems.  For example,
one of Elkin's rules is that ``Quips, jokes, or sarcastic comments,
even if kindly meant, are inappropriate.''  Rather than forbidding
humour, it may be better for individual comments to be rated as
helpful or non-helpful.  Again, in the first instance, usefulness
and interest might be judged in terms of explicit criteria for serendipity;
see \cite{corneli15cc,pease2013discussion}.
The key criterion in this regard is the \emph{focus shift}. 
This is the creation of a novel problem, comprising the move
from discovery of interesting data to the invention of an application.
This process is distinct from identifying routine errors in a written work.  Nevertheless, from a
computational standpoint, noticing and being robust to certain kinds
of errors is often a preliminary step.  For example, the work might
contain a typo, grammatical or semantic error, while being logically
sound.  In a programming setting, this sort of problem can lead to
crashing code, or silent failure.  In general communicative context,
argumentation may be logically sound, but not practically useful or
poorly exposited.  Finally, even a masterful, correct, and fully
spellchecked piece of argumentation may not invite further dialogue,
and so may fail to open itself to further learning.  Identifying and
engaging with this sort of deeper issue is something that skillful
workshop participants may be able to do.  Dialogue in the workshop
can build on strong or less strong work -- but provoking interpretative
thoughts and comments always require a thoughtful critical presence and
the ability to engage.  This can be difficult for humans and poses a range
of challenges for computers -- but also promises some interesting
results.

\bigskip

\section{Related work}\label{sec:ww-related}


In considering the potential and contribution of the Writers Workshop model outlined in Section \ref{sec:writers-workshop}, we posit that the Writers Workshop model is useful for encouraging feedback in computational systems, and in particular systems that are designed to be creative or serendipitous. 

Feedback has long been a central concept in AI-related fields, particularly cybernetics \cite{pickering2002cybernetics}, as well as related disciplines, like neuroscience \cite{ashby1952,seth2015cybernetic}.  The basic types of feedback are \emph{positive}, and \emph{negative}: positive feedback increases perturbation, whereas negative feedback reduces perturbation.  Sources of negative feedback are thereby useful in system control applications (and their analogues).   Feedback and control (including feedback about control) is understood to be relevant to thinking about \emph{learning} and \emph{communication} \cite{bateson+steps+2000}.  We focus on the role that communicative feedback can play in computational creativity and computational serendipity and discuss previous related work on incorporating feedback into such computational systems.

\subsection{Feedback in computational creativity} \label{ref:related-computational-creativity}

\emph{Creativity} is often envisaged as involving cyclical processes
(e.g.~Dickie's \citeyear{dickie1984art} art circle, Pease and Colton's
\citeyear{pease2011computational} Iterative
Development-Expression-Appreciation model).  There are opportunities
for embedded feedback at each step, and the creative process itself is ``akin
to'' a feedback loop.  However, despite these strong intimations of
the central importance of feedback in the creative process, our sense
is that feedback has not been given a central place in research on
computational creativity.  In particular, current systems in
computational creativity, almost as a rule, do \emph{not} consume or
evaluate the work of other systems.\footnote{An exception to
  the rule is Mike Cook's {\sf AppreciationBot}
  (\url{https://twitter.com/AppreciationBot}), which is a reactive
  automaton that ``appreciates'' tweets from {\sf MuseumBot}.}

\citeA{gervas2014reading} theorise a creative cycle of narrative
development as involving a Composer and an Interpreter, in such a way
that the Composer has internalised the interpretation functionality.
Individual creativity is not the poor relation of social creativity,
but its mirror image.  Nevertheless, even when computer models
explicitly involve multiple agents and simulate social creativity
\cite<like>{saunders2001digital}, they rarely make the jump to involve
multiple systems.  The ``air gap'' between computationally creative
systems is very different from the historical situation in human
creativity, in which different creators and indeed different cultural
domains interact vigorously \cite{geertz1973interpretation}.

\subsection{Feedback in computational serendipity} \label{ref:related-computational-serendipity}

The term computational serendipity is rather new, but its foundations
are well established in prior research.  

\citeA{grace2014using} examine \emph{surprise}
in computing, seeking to ``adopt methods from the
field of computational creativity [$\ldots$] to the generation of
scientific hypotheses.''  This is an example of an effort focused on
computational \emph{invention}. 

An area of AI where serendipity can be argued to play an important part is 
in pattern matching. Current computer programs are able to identify known patterns and
``close matches'' in data sets from certain domains, like music
\cite{meredith2002algorithms}.  Identifying known patterns is a
special case of the more general concept of \emph{pattern mining}
\cite{bergeron2007representation}.  In particular, the ability to
extract \emph{new} higher order patterns that describe exceptions is
an example of ``learning from feedback.''  Deep learning and
evolutionary models increasingly use this sort of idea to facilitate
strategic discovery \cite{samothrakis2011approximating}.  Similar
ideas are considered in business applications under the heading
``process mining'' \cite{van2011process}.

In earlier work \cite{corneli15cc,corneli15iccc}, we used the idea of
dialogue in a Writers Workshop framework to sketch a ``theory of
poetics rooted in the making of boundary-crossing objects and
processes'' and described (at a schematic level) ``a system that can
(sometimes) make `highly serendipitous' creative advances in computer
poetry'' while ``drawing attention to theoretical questions related to
program design in an autonomous programming context.''

\subsection{Communications and feedback}

The Writers Workshop heavily relies on communication of feedback within the workshop.
 Gordon Pask's conversation theory, reviewed in
\cite{conversation-theory-review,boyd2004conversation}, goes
considerably beyond the simple process language of the workshop,
although there are structural parallels.  We see that a basic Pask-style
learning conversation bears many similarities to the Writers Workshop model of communicative feedback
\cite[p. 190]{boyd2004conversation}: 

\begin{center}
\begin{fminipage}{.8\columnwidth}
\begin{minipage}{1\textwidth}
\begin{enumerate}[itemsep=0pt,rightmargin=10pt]
\item Conversational participants are carrying
out some actions and observations;
\item Naming and recording what action is being done;
\item Asking and explaining why it works the way
it does;
\item Carrying out higher-order methodological discussion; and, 
\item Trying to figure out why unexpected results occurred.
\end{enumerate}
\end{minipage}
\end{fminipage}
\end{center}

Variations to the underlying system, protocol, and the schedule of
events should be considered depending on the needs and interests of
participants, and several variants can be tried.  On a pragmatic
basis, if the workshop proved quite useful to participants, it could
be revised to run monthly, weekly, or continuously.\footnote{For a
  comparison case in computer Go, see
  \url{http://cgos.computergo.org/}.}

\section{Case study: Flowcharts and Feedback} \label{sec:implementation}

This section describes work that is currently underway to implement the
Writers Workshop model, not only within one system but as a new paradigm
for collaboration among disparate projects.  In order to
bring in other participants, we need a neutral environment that is not
hard to develop for: the \Fw\ system mentioned in Section
\ref{sec:ww-model} offers one such possibility.  The basic primary
objects in the \Fw\ system are \emph{flowcharts}, which are comprised of
interconnected \emph{process nodes}
\cite{charnley2014flowr,colton-flowcharting}.  Process nodes specify
input and output types, and internal processing can be implemented in
Java, or other languages that interoperate with the JVM, or by
invoking external web services.  One of the common applications to
date is to generate computer poetry, and we will focus on that domain
here.

A basic set of questions, relative to this system's components, are as 
follow:
\begin{enumerate}
\item \emph{Population of nodes}: What can they do?  What do we learn
  when a new node is added?
\item \emph{Population of flowcharts}: \citeA{pease2013discussion}
  have described the potentially-serendipitous repair of ``broken''
  flowcharts when new nodes become available; this suggests the need for
  test-driven development framework.
\item \emph{Population of output texts}: How to assess and comment on
  a generated poetic artefact?
\end{enumerate}

In a further evolution of the system, the sequence of steps in a
Writers Workshop could itself be spelled out as a flowchart.  The
process of reading a poem could be conceptualised as generating a
semantic graph \cite{harrington2007asknet,francisco2006automated}.
Feedback could be modelled as annotations to a text, including
suggested edits.  These markup directives could themselves be
expressed as flowcharts.  A standardised set of markup structures may
partially obviate the need for strong natural language understanding,
at least in interagent communication.  For example, we might agree
that {\tt observations} will consist of stand-off annotations that
connect individual textual passages to public URIs using a limited
comparison vocabulary, and that {\tt suggestions} will consist of
simple stand-off line-edits, which may themselves be marked up with
rationale.  In fact, we would not need to be so restrictive: it would
not be much more complicated to add annotations to the annotations, so
as to be able to form composite statements that express the
relationships between different pieces of the annotated texts.
Whatever restrictions we ultimately impose on annotation formats, as
well as similar restrictions around constrained turn-taking in the
workshop, could be progressively widened in future versions of the
system.
The way the poems that are generated, the models of poems that are
created, and the way the feedback is generated, all depend on the
contributing system's body of code and prior experience, which may
vary widely between participating systems.  In the list \ref{step:model}--\ref{step:update} of functional steps below, all of
the functions could have a subscripted ``$\mathcal{E}$'', which is
omitted throughout.  Exchanging path dependent points of view will tend to produce results
that are different from what the individual participating systems would
have come up with on their own.

\begin{enumerate}[label=\Roman*.]
\item\label{step:model} Both the author and critic should be able to work with a model
  of the text.  Some of the text's features may be explicitly tagged
  as ``interesting.''  Outstanding questions may possibly be
  brought to the attention of critical listeners, e.g.~with the
  request to compare two different versions of the poem ({\tt
    presentation}, {\tt listening}).
\begin{enumerate}[label=\arabic*.]
\item \emph{A model of the text}. $m: T\rightarrow M$.
\item \emph{Tagging elements of interest}. $\mu: M\rightarrow I$.
\end{enumerate}
\item\label{step:feedback} Drawing on its experience, the critic will use its model of the
  poem to formulate feedback ({\tt feedback}).
\begin{enumerate}[label=\arabic*.]
\item \emph{Generating feedback}. $f: (T,M,I)\rightarrow F$.
\end{enumerate}
\item\label{step:rationale} Given the constrained framework for feedback, statements about
  the text will be straightforward to understand, but rationale for
  making these statements may be more involved ({\tt questions}, {\tt
    replies}).
\begin{enumerate}[label=\arabic*.]
\item \emph{Asking for more information}. $q: (M,F,I) \rightarrow Q$.
\item \emph{Generating rationale}. $a: (M,F,Q) \rightarrow \Delta F$.
\end{enumerate}
\item\label{step:update} Finally, feedback may affect the author's model of the world, and the way future poems are generated ({\tt reflection}).
\begin{enumerate}[label=\arabic*.]
\item \emph{Updating point of view}. $\rho: (M,F) \rightarrow \Delta\mathcal{E}$.
\end{enumerate}
\end{enumerate}

The final step is perhaps the most interesting one, since invites us
to consider how individual elements of feedback can ``snowball'' and
go beyond line-edits to a specific poem to much more fundamental
changes in the way the presenting agent writes poetry.  Here methods
for pattern mining, discussed in Section
\ref{ref:related-computational-serendipity}, are particularly relevant.
If systems can share code (as in our sample dialogue in Section
\ref{sec:dialogue-example}) this will help with the
rationale-generating step, and may also facilitate direct updates to
the codebase.  However, shared code may be more suitably placed into
the common pool of resources available to \Fw\ than copied over as
new ``intrinsic'' features of an agent.

Although different systems with different approaches and histories are
important for producing unexpected effects, ``offline'' programmatic
access to a shared pool of nodes and existing flowcharts may be
useful.  Outside of the workshop itself, agents may work to recombine
nodes based on their input and output properties to assemble new
flowcharts.  This can potentially help evaluate and evolve the
population of nodes programmatically, if we can use this sort of
feedback to define fitness functions.  The role of
temporality is interesting: if the workshop takes place in real time,
this will require different approaches to composition that takes place
offline \cite{perez2013rolling}.
Complementing these ``macro-level'' considerations, it is also worth
commenting on the potential role of ``micro-level'' feedback within
flowcharts.  Local evaluation of output from a predecessor node could
feed backwards through the flowchart, similar to backpropagation in
neural networks.  This would rely on a reduced version of the
functional schema described above.

\section{Concluding discussion and future directions}

 We have described a \emph{general} and \emph{computationally
   feasible} model for using feedback in AI systems, particularly
 creative systems.  The Writers Workshop concept, borrowed from
 creative writing, is transformed into a model of a structured
 approach to eliciting, processing and learning from feedback.  To
 better evaluate how the Writers Workshop model helps us advance in
 our goal of incorporating feedback into artificial creativity, we
 critically considered how the model fits into related work. In
 particular, we found that serendipity, a key concept within
 creativity and AI more generally, is a concept with which the Writers
 Workshop model could assist computational progress.
In this respect, we should highlight the difference between
``global'' analytics describing the collection of nodes and
flowcharts in the \Fw\ ecosystem, and the path-dependent
process of analysis and synthesis that takes place in a workshop setting.

Our preliminary implementation work (Section \ref{sec:implementation}) shows
that the model can be transferred to a functional implementation.  This
work highlights several considerations relevant to further work with
the Writers Workshop model:

\begin{itemize}
\item Each contributing system should come to the workshop with at
  least a basic awareness of the workshop protocol, with work to
  share, and prepared to give constructive feedback to other systems.
\item The workshop itself needs to be prepared, with a suitable
  communication platform and a moderator or global flowchart for
  moving the discussion from step to step.
 \item A controlled vocabulary for communications and interaction
   would be a worthwhile pursuit of future research, perhaps based on
   an ontology inspired by the Interaction Network
   Ontology.\footnote{The Interaction Network Ontology primarily describes
     interactions within humans as opposed to within human societies;
     a distinct \emph{Social} Interaction Ontology does not seem to
     exist at present.  However, the classes of the Interaction Network Ontology appear to be quite broadly relevant.
     This ontology is documented
     at \url{http://www.ontobee.org/browser/index.php?o=INO}.  Its URI
     is
     \url{http://svn.code.sf.net/p/ino/code/trunk/src/ontology/INO.owl}.}
\item In order to get the most value out of the workshop experience,
  systems (and their wranglers) should ideally have questions they are
  investigating.  As discussed above, prior experience plays an
  important role in every step.  This opens up a range of issues for
  further research on modelling motivations and learning from
  experience.
\item Systems should be prepared to give feedback, and to carry out
  evaluations of the helpfulness (or not) of feedback from other
  systems and of the experience overall.
\end{itemize}
  
  Developing systems that could successfully navigate this
  collaborative exercise would be a significant advance in the field
  of computational creativity.  Since the experience is about learning
  rather than winning, there is little motivation to ``game the
  system'' (cf. \citeNP{lenat1983eurisko}). Instead the emphasis is
  squarely upon mutual benefit: computational systems helping to
  develop each other through communication and feedback.


The benefits of the Writers Workshop approach could innovate well beyond models for 
feedback and communication within a particular environment or restricted domain. 
Following the example of the Pattern Languages of Programming (PLoP) community, we propose that the Writers Workshop model could be deployed
within the Computational Creativity community to design a workshop in
which the participants are computer systems instead of human authors.
The annual International Conference on Computational Creativity
(ICCC), now entering its sixth year, could be a suitable venue. 

Rather than the system's creator presenting the system in a
traditional slideshow and discussion, or a system ``Show and Tell,''
the systems would be brought to the workshop and would present their
own work to an audience of other systems, in a Writers Workshop
format.  This could be accompanied by a short paper for the conference
proceedings written by the system's designer describing the system's
current capabilities and goals.  
If the workshop model really works well, future publications might adapt to include
traces of workshop interactions, commentary from a system on
other systems, and offline reflections on what the system might change
about its own work based on the feedback it receives.  Paralleling the PLoP
community, it could become standard to incorporate the workshop
into the process of peer review for the new \emph{Journal of
  Computational Creativity}.\footnote{\url{http://www.journalofcomputationalcreativity.cc}} AI systems that review each other would surely be a major demonstration and acknowledgement of the usefulness of feedback within AI.

In closing, we wish to return briefly to the scenario of computer
generated feedback in educational contexts that we raised at the
beginning of this paper and then set aside.  The elements of our
functional design for sharing feedback among computational agents has
a range of features that continue to be relevant for generating useful
feedback with human learners.  Students are experience-bound, and a
robust approach to formative assessment and feedback should take into
account the student's historical experience, so far as this can be
known or inferred.  In order for feedback, recommendations, and so on to
adequately take individual history into account, sophisticated modelling and
reasoning would be required.
Nevertheless, from the point of view of participating computational
agents, a student may simply look like another agent.  It is particularly in this
regard that computational models of learning from feedback are seen
as fundamental.

\section*{Acknowledgement}
Joseph Corneli's work on this paper was supported by the Future and Emerging
Technologies (FET) programme within the Seventh Framework Programme
for Research of the European Commission, under FET-Open Grant number
611553 (COINVENT).

\bibliographystyle{apacite}
\bibliography{./biblio}

\end{document}